\documentclass[10pt,twocolumn,letterpaper]{article}

\usepackage[pagenumbers]{cvpr} % To force page numbers, e.g. for an arXiv version

\definecolor{cvprblue}{rgb}{0.21,0.49,0.74}
\usepackage[pagebackref,breaklinks,colorlinks,allcolors=cvprblue]{hyperref}

\def\paperID{5614} % *** Enter the Paper ID here
\def\confName{CVPR}
\def\confYear{2025}

\title{A General Purpose Spectral Foundational Model for Both Proximal and Remote Sensing Spectral Imaging}

\author{William Michael Laprade\textsuperscript{1}
\and
Jesper Cairo Westergaard\textsuperscript{3}
\and
Svend Christensen\textsuperscript{3}
\and
Mads Nielsen\textsuperscript{2}
\and
Anders Bjorholm Dahl\textsuperscript{1}
\and
\textsuperscript{1}Department of Applied Mathematics and Computer Science, Technical University of Denmark\\
\textsuperscript{2}Department of Computer Science, University of Copenhagen\\
\textsuperscript{3}Department of Plant and Environmental Sciences, University of Copenhagen}

\begin{document}
\maketitle
\begin{abstract}
Spectral imaging data acquired via multispectral and hyperspectral cameras can have hundreds of channels, where each channel records the reflectance at a specific wavelength and bandwidth. Time and resource constraints limit our ability to collect large spectral datasets, making it difficult to build and train predictive models from scratch. In the RGB domain, we can often alleviate some of the limitations of smaller datasets by using pretrained foundational models as a starting point. However, most existing foundation models are pretrained on large datasets of 3-channel RGB images, severely limiting their effectiveness when used with spectral imaging data. The few spectral foundation models that do exist usually have one of two limitations: (1) they are built and trained only on remote sensing data limiting their application in proximal spectral imaging, (2) they utilize the more widely available multispectral imaging datasets with less than 15 channels restricting their use with hundred-channel hyperspectral images. To alleviate these issues, we propose a large-scale foundational model and dataset built upon the masked autoencoder architecture that takes advantage of spectral channel encoding, spatial-spectral masking and ImageNet pretraining for an adaptable and robust model for downstream spectral imaging tasks. Data and code is available here: [URL to come]
\end{abstract}    
\begin{figure}[ht]
     \centering
     \includegraphics[width=0.99\linewidth]{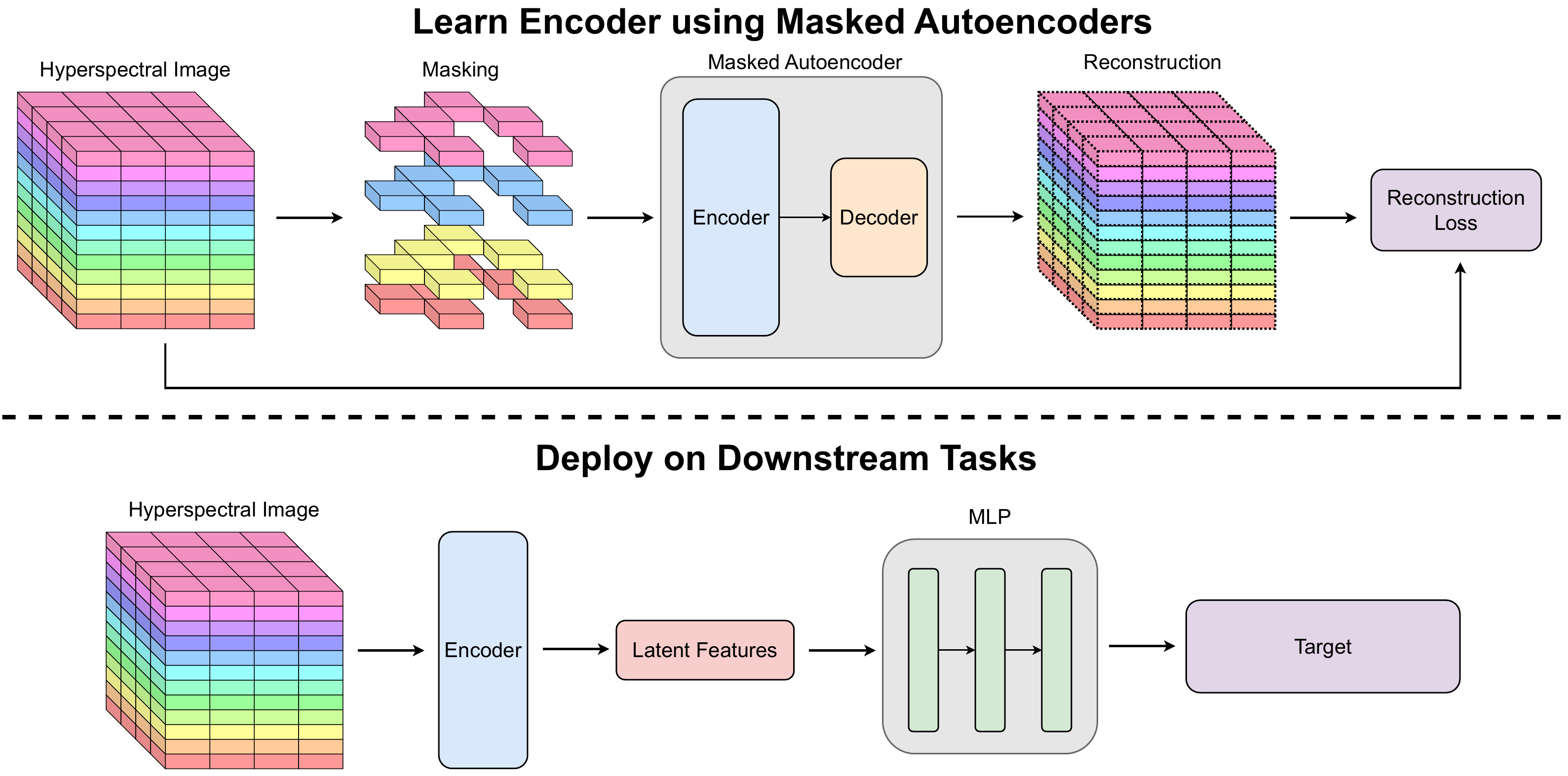}
     \caption{A masked autoencoder is used to pretrain a ViT encoder using hyperspectral data. The encoder can then be used for finetuning on a wide range of hyperspectral tasks across both proximal and remote sensing imagery.}
     \label{fig:overview}
\end{figure}

\section{Introduction}
\label{sec:intro}

Spectral imaging captures, for each pixel, spectral reflectance at specific wavelengths in the electromagnetic spectrum. It produces an image, where each channel contains reflectance information at a specific wavelength and bandwidth. The combination of spectral and spatial information in a hyperspectral image gives unique opportunities to learn powerful deep discriminative models. Hyperspectral remote sensing images are available in great numbers from satellites such as EnMap \citep{Guanter2015,Storch2023}, whereas proximal data, \ie images taken at a close distance, is limited, and proximal datasets are typically small. 

To leverage the power of deep learning for small proximal hyperspectral datasets, we suggest training foundational models. Available datasets come with a different number of spectral channels, but to ensure sufficient variation in the training data, it is essential to include all available data. Therefore, we suggest a transformer-based masked auto-encoder model that utilizes positional encoding of the spectral channels. Spectral encoding makes the model agnostic to the wavelengths of the spectral channels and the transformer architecture makes our model agnostic to the number of input channels. Our model is illustrated in \cref{fig:overview}.

Multispectral imaging devices typically record only a few channels at very specific wavelengths and bandwidths while hyperspectral imaging cameras record a high resolution spectrum for each pixel. In contrast to RGB imaging which produces images designed to simulate what the human eye sees, spectral imaging records detailed and specific reflectance information that can be used to identify and classify materials and objects beyond the capabilities of a typical RGB image \cite{shippert2004use}.

Applications for spectral imaging can be divided into two categories: (1) remote sensing -- distant imaging from satellite or drone/aircraft cameras, (2) proximal imaging -- imaging at distances similar to what humans see visually day-to-day. In remote sensing, spectral imaging sees use in \eg climate research \cite{calin2021application}, forestry \cite{adao2017hyperspectral}, geography \cite{song2017deep}, agriculture research \cite{lu2020recent}, astronomy \cite{hege2004hyperspectral}, while proximal spectral imaging finds applications in waste sorting \cite{zheng2018discrimination}, counterfeit detection \cite{huang2022recent}, medical diagnostics \cite{lu2014medical}, and others \cite{makki2017survey}. Such a wide range of applications of spectral imaging in both proximal and remote sensing settings, combined with the limited availability of data hint at the need for a large pretrained spectral foundational model to use a pretrained base for training models on these small datasets.

One of the reasons building such a model has proven difficult is due to how spectral data is often collected. Spectral cameras can be expensive and the imaging time-consuming. Publicly available datasets are limited in scale and often assigned to a very specific application (\eg analyzing disease among a specific variety of wheat). Due to this specificity, large pretraining datasets, in proximal imaging especially, are rare or non-existing. To deal with this, we have built a large pretrainined dataset with 16,529 hyperspectral images compiled from a number of existing datasets as well as a few datasets that we collected on our own.

In combination with our dataset, we have designed a masking method that ensures both spatial and spectral knowledge is learned and built a spectral encoding method that uses wavelength and bandwidth information to build a model that is robust to both existing and new spectral imaging systems.

\subsection{Related work}

In fields where data is limited, pretraining often becomes an essential first step in building deep learning models. In computer vision, self-supervised methods are the backbone of most of these foundational models. Utilizing large datasets such as ImageNet \cite{deng2009imagenet}, self-supervision methods work by applying either a masked reconstruction \cite{haresamudram2020masked} or contrastive learning \cite{jaiswal2020survey} based approach. Contrastive learning approaches, used models like DINO \cite{caron2021emerging} and MoCo \cite{he2020momentum}, use loss functions that compare the similarity of feature representations of original and augmented images, while reconstruction-based approaches such as MAE \cite{he2022masked} and VAE \cite{he2020momentum} attempt to reconstruct images from augmented or masked inputs. 

In this paper, we use the masked autoencoders (MAE) architecture to build meaningful representations for downstream tasks. The architecture is built upon a ViT \cite{dosovitskiy2020image} encoder and a much smaller ViT decoder. Images are patched and then a portion of the patches are removed prior to being fed into the encoder. The decoder then attempts to reconstruct the missing patches. The original paper \cite{he2022masked} found that removing 75\% of patches was the optimal ratio for pretraining the ViT encoder on RGB data. When utilized for video, where the data is spatially redundant along the temporal axis, a ratio of 90\% was found to be optimal \cite{tong2022videomae,feichtenhofer2022masked}. In the case of spectral imaging, which we would argue has a higher redundancy along the spectral axis than video does along the temporal axis, also often utilizes a 90\% masking ratio \cite{braham2024spectralearth,li2024s2mae}.

There exist some spectral foundation models such as SpectralGPT \cite{hong2024spectralgpt}, SS2MAE \cite{li2024s2mae}, HyperSIGMA \cite{wang2024hypersigma} and SpectralEarth \cite{braham2024spectralearth}. SpectralGPT and SS2MAE are trained primarily on large datasets of small Sentinel-2 imaging patches with 12 or 13 channels and are limited to applications involving multispectral imaging. For hyperspectral data, SpectralEarth and HyperSIGMA utilize 224-channel EnMap \cite{Storch2023,Guanter2015} satellite imaging to build powerful foundational models for hyperspectral imaging. None of the existing spectral foundational models utilize large-scale datasets of proximal imaging. The lack of research into this area is the primary motivation for this paper. 

There exist many applications which utilize hyperspectral imaging for proximal image analysis. It has seen uses in everything ranging from skin cancer detection \cite{huang2023review} and food classification \cite{saha2021machine}, to forgery \cite{silva2014near} and landmine detection \cite{makki2017survey}. In agricultural research, for example, hyperspectral imaging has advantages over RGB and even multispectral imagery. By finding specific narrow bands of interest vegetation indices can be computed that can be used to detect specific biological signals. These vegetation indices have been used to distinguish invasive species in a dune ecosystem \cite{GroeStoltenberg2016}, to discern symptoms of multiple diseases in sugar beets \cite{Mahlein2012}, and for detecting mineral deficiencies in maize leaves \cite{Song2024}. 

Despite a wide range of applications for hyperspectral imaging, most of the datasets are fairly small, due to the difficulties in setting up and using hyperspectral imaging systems. With limited data, it is difficult to train high-performing but data-hungry architectures like vision transformers.
\section{Data}
\label{sec:data}

The dataset we have compiled consists of 16,518 hyperspectral images across 7 pre-existing hyperspectral datasets and 3 datasets collected using our SpecimIQ hyperspectral camera and 1 remote sensing dataset built from EnMAP satellite hyperspectral images. See \cref{tab:datasets} for details.

\subsection{Existing proximal datasets}
\begin{figure*}[ht]
     \centering
     \includegraphics[width=0.99\linewidth]{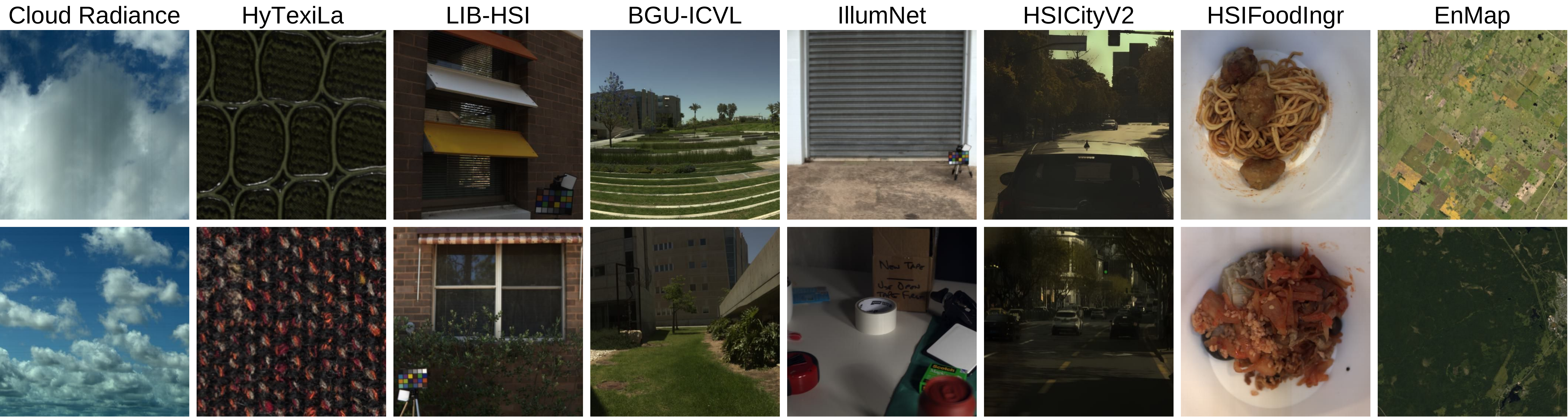}
     \caption{Sample images from existing datasets in RGB.}
     \label{fig:compiled}
\end{figure*}

There is limited access to high-quality hyperspectral data, and many datasets are only released in downscaled versions of the original data. From the limited remaining options, we selected 7 datasets to include in our pretraining dataset. Example images are shown in \cref{fig:compiled}. These are chosen to get a wide range of imaged objects and cameras. The compiled datasets include images ranging from car-mounted city views to clouds and cooked dishes.

\subsection{Collected proximal datasets}
\begin{figure}[ht]
     \centering
     \includegraphics[width=0.99\linewidth]{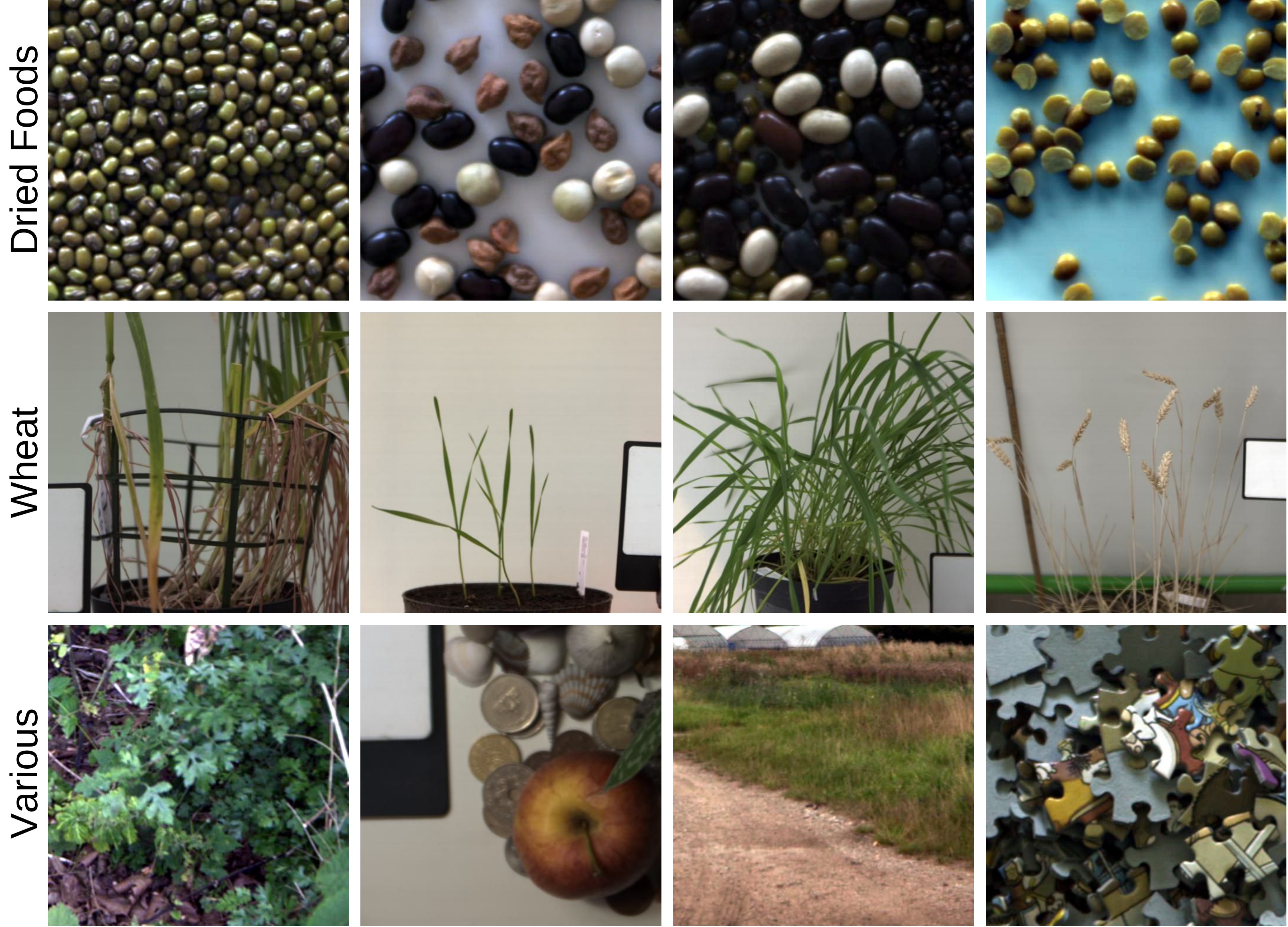}
     \caption{Sample images from collected datasets in RGB.}
     \label{fig:collected}
\end{figure}

Using a portable SpecimIQ \cite{behmann2018specim} hyperspectral imaging camera, we collected 3 additional datasets to include as part of our large pretraining dataset. These include a general imaging dataset of areas around Copenhagen, Denmark, a wheat growth dataset collected from a greenhouse that includes growth week and soil type labels, and a dried foods dataset containing 18 different types of lentils, beans, and quinoa with associated labels and mixtures. The SpecimIQ camera records images with $512 \times 512$ spatial resolution and 204 spectral channels and depending on the dataset patched into smaller $256 \times 256$ images. Example images are shown in \cref{fig:collected}.

\noindent\textbf{General imaging dataset:} We collected 135 images around Copenhagen, Denmark in both Winter and Summer using a SpecimIQ hyperspectral camera. Each image is originally $512 \times 512$ but cropped to $256 \times 256$ patches for a total of 540 images with 204 spectral channels and includes a variety of objects from both outdoors and indoors. 

\noindent\textbf{Dried foods dataset:} This is a multi-label classification dataset that includes 384 images of 18 different dried foods. These include 6 different types of lentils, 6 different beans, 3 types of chickpeas, and 3 types of quinoa. Each type is imaged twice at different densities. The dataset also includes 12 mixtures of 3 different types and 12 mixtures of 6 different types. The dataset is divided into 192 training images, 96 validation images, and 96 test images. 

\noindent\textbf{Wheat dataset:} A second multi-label classification dataset was collected from a greenhouse trial at the University of Copenhagen's Taastrup Campus. It contains 156 images of the cereal winter wheat grown in two soil types: sandy soil and peat moss soil. Every other week, a new triplet of pots was sown for each treatment. The images show wheat plants in different growth stages, from young tillering plants to fully senesced, and include 18 classification labels for plant age and soil types. Images are white-tile calibrated at the time of acquisition to obtain true reflectance. The dataset has 62 training images, 47 validation images, and 47 test images.

\subsection{EnMap remote sensing}

In additional to proximal imaging data, we also extracted 8,504 patches from the EnMap \cite{Storch2023,Guanter2015} hyperspectral imaging data source with 224 spectral bands. In order to obtain a wide variety of locations and times of year, we divided the available EnMap data into 30 equally spaced latitude sections, 60 equally spaced longitudinal sections, 4 seasons, and 15 biome types. For each of the 108,000 possible combinations, if that combination has existing EnMap tiles, we select one tile at random to include in the dataset. This resulted in 2,126 tiles which are then split into 4 equally sized patches.

\begin{table*}[ht]
\centering
\begin{tabular}{@{}rrcccccrr@{}}
\toprule
Dataset & \# Images & Width & Height & Channels & Wavelengths & Bandwidths & Spatial size & Total Size \\ \midrule
\textbf{DriedFoods} & 384 & 256 & 256 & 204 & 397 - 1003 & 7.00 & 25.2 & 5,141 \\
CloudRadiance \cite{CloudRadiance} & 30 & 672 & 672 & 462 & 388 - 1004 & 1.90 & 13.5 & 6,237 \\
\textbf{Various} & 540 & 256 & 256 & 204 & 397 - 1003 & 7.00 & 35.4 & 7,222 \\
\textbf{Wheat} & 156 & 512 & 512 & 204 & 397 - 1003 & 7.00 & 40.9 & 8,342 \\
HyTexiLa \cite{khan2018hytexila} & 448 & 512 & 512 & 186 & 405 - 995 & 3.19 & 117.4 & 21,836 \\
LIB-HSI \cite{habili2022hyperspectral} & 513 & 512 & 512 & 204 & 397 - 1003 & 7.00 & 134.5 & 27,438 \\
BGU-ICVL \cite{arad2016sparse} & 201 & 622 & 672 & 519 & 390 - 1040 & 1.20 - 1.32 & 84.0 & 43,596 \\
IllumNet \cite{habili2023automatic} & 1,024 & 512 & 512 & 204 & 397 - 1003 & 7.00 & 269.5 & 54,978 \\
HSICityV2 \cite{Shen2023} & 1,330 & 893 & 672 & 128 & 449 - 955 & 5.00 & 798.1 & 102,157 \\
HSIFoodIngr \cite{xia2023hsifoodingr} & 3,388 & 512 & 512 & 204 & 397 - 1003 & 7.00 & 888.1 & 181,172 \\
EnMap \cite{Storch2023,Guanter2015} & 8,504 & 444 & 447 & 218 & 418 - 2445 & 5.80 - 11.30 & 1687.2 & 367,810 \\ \bottomrule
\end{tabular}
\caption{Breakdown of our compiled dataset. Collected datasets in bold. Spatial size measures number of megapixels spatially. Total size is a measure of the total number values (spatial size $\times$ number of channels). The datasets that we have collected are marked in bold and available from: [URL to come].}
\label{tab:datasets}
\end{table*}
\section{Methods}

\begin{figure*}[ht]
     \centering
     \includegraphics[width=0.99\linewidth]{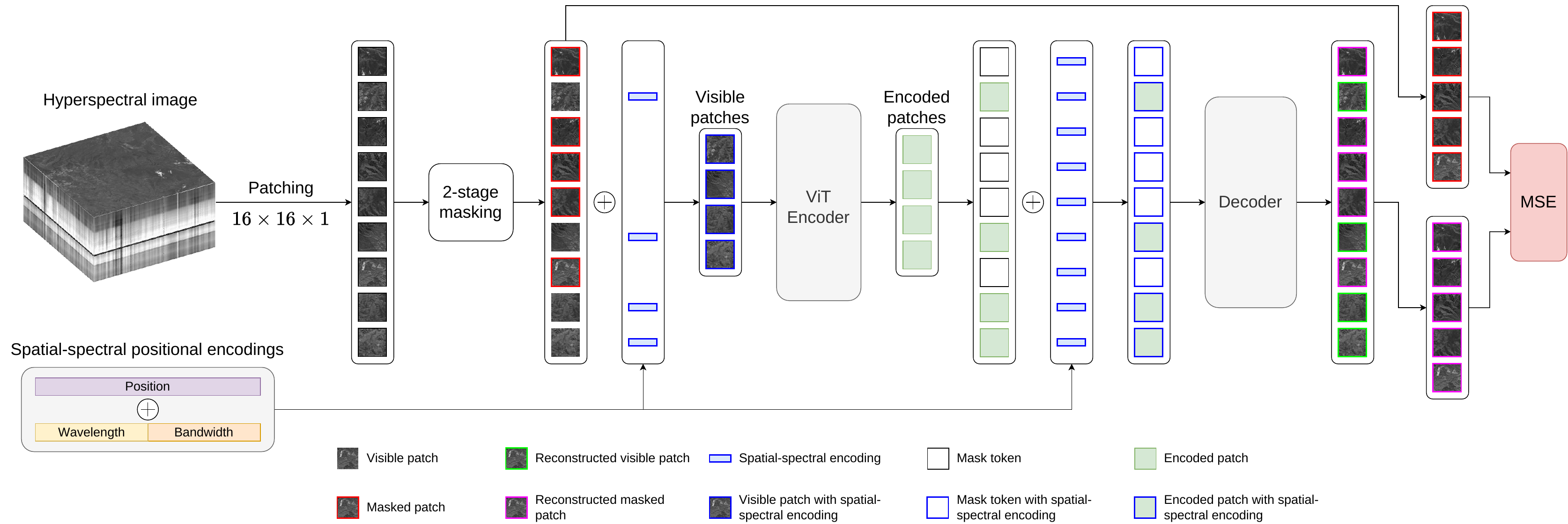}
     \caption{Our MAE architecture. The hyperspectral image is divide into single channel $16 \times 16 \times 1$ patches. Masking removes a majority of patches and the remaining patches are fed into the encoder. Utilizing the encoded information from the visible patches the decoder attempts to reconstruct the masked patches. MSE loss is computed only on the reconstructed masked patches.}
     \label{fig:mae}
\end{figure*}

We build upon the widely used MAE architecture and adapt it to hyperspectral data with two major changes: (1) Our implementation uses a unique 3D masking strategy, that ensures both spatial and spectral information is learned by ensuring that both entire spatial regions and entire spectral channels are removed. (2) We introduce a spatial-spectral positional encoding that utilizes the both the spatial coordinates and the wavelength and bandwidth information from the camera. The architecture of our MAE model is shown in \cref{fig:mae}.

\subsection{Channel sampling}
Random channel selection is also applied during hyperspectral pretraining to select 100 channels at random from each image to pass into the model. This is done to both reduce the computation time over including all channels and to simulate various camera sensors that might have a different wavelength and bandwidth coverage than what is included in our dataset. It is equivalent to hierarchical channel sampling \cite{bao2024channelvisiontransformersimage} with a fixed $m = 100$. Due to significant spectral redundancy in hyperspectral imaging and the additional dropout incurred in MAE self-supervision, we expect that having a non-fixed $m$ is not neccessary for pre-training. Although further experimentation is required. The final input is a $224 \times 224 \times 100$ image.

\subsection{Patching}
We divide each hyperspectral image with shape $W \times H \times C$, into patches of shape $16 \times 16 \times 1$. Despite the significant increase in sequence length that single channel patches creates, it has been suggested in \cite{bao2024channelvisiontransformersimage} that per-channel tokens with dropout improve performance at test-time with varying channel counts. 

\subsection{Masking}
\begin{figure}[ht]
     \centering
     \begin{subfigure}[b]{0.15\textwidth}
         \centering
         \includegraphics[width=\textwidth]{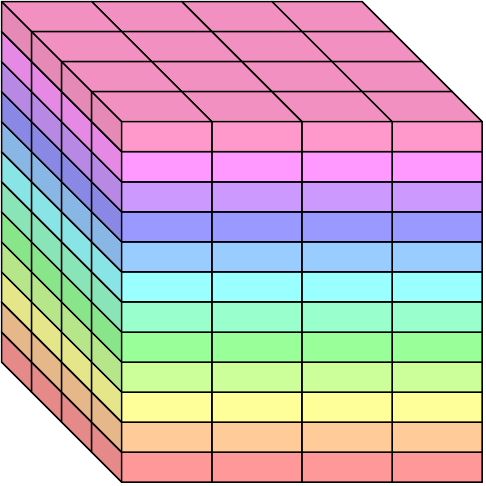}
         \caption{Input}
         \label{fig:mask_hsi}
     \end{subfigure}\hfill
     \begin{subfigure}[b]{0.13\textwidth}
         \centering
         \includegraphics[width=\textwidth]{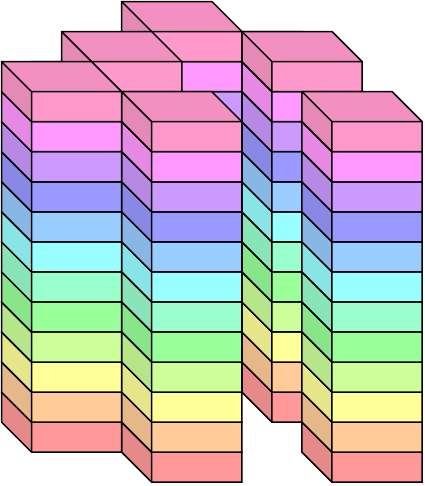}
         \caption{Spatial}
         \label{fig:mask_spatial}
     \end{subfigure}\hfill
     \begin{subfigure}[b]{0.13\textwidth}
         \centering
         \includegraphics[width=\textwidth]{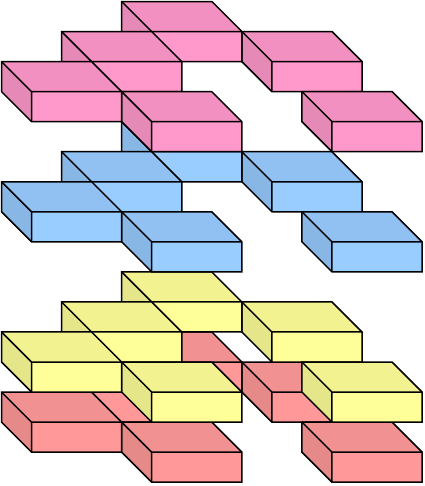}
         \caption{Spatial-spectral}
         \label{fig:mask_spatial_spectral}
     \end{subfigure}\hfill
     \caption{Two-stage masking. An initial column masking to remove entire spatial regions from the image followed by a channel masking to remove spectral information.}
     \label{fig:mask}
\end{figure}

Instead of applying random 3D masking as most MAE implementations do, we opt for a two-stage masking strategy to ensure the model learns both spatial and spectral information effectively as illustrated in \cref{fig:mask}. In the first stage, we apply random tube masking along the spectral dimension as shown in \cref{fig:mask_spatial} with a ratio of 75\%. This ensures that entire spatial regions are deleted, thus forcing the model to learn how to reconstruct spatial features effectively. In the second stage, shown in \cref{fig:mask_spatial_spectral}, we apply random channel masking to delete 65-95\% of the channels from the image ensuring the model also learns spectral information effectively. After both masking stages, our spatial-spectral masking results in a final masking ratio between 91.25\% and 98.75\% depending on the channel masking ratio.

\subsection{Spatial-spectral encoding}
Prior to being fed into the MAE, the patches are flattened and scaled to 1024 via a fully connected layer. A spatial-spectral embedding is then added to this 1024-length vector. The spatial-spectral embedding is the summation of two embeddings $S_{\text{spatial}}$ and $S_{\text{spectral}}$ as opposed to the simpler 3D positional encoding. This ensures that the spectral and spatial information is treated separately by the model. We define $S_{\text{spatial}}$ as a learned feature vector instead of using sinusoidal embeddings. $S_{\text{spectral}}$ is the concatenation of the output of two linear 512-length layers $L_{\lambda}$ and $L_{B}$. The input to these two layers is the scaled wavelength, $\lambda$, and bandwidth, $B$, values for each channel. Hyperspectral cameras typically provide the wavelength and bandwidth (FWHM) of each recorded channel. Most hyperspectral imaging data is recorded in the 400-2500nm range and has bandwidths between 1-250nm. Prior to being passed into the $L_{\lambda}$ and $L_{B}$ layers, $\lambda$ and $B$ are scaled by factor of $\frac{1}{2500}$ and $\frac{1}{250}$ for the wavelength and bandwidth, respectively.

\subsection{Training procedure}

Similar to many architectures, we utilize progressive pretraining to gradually tune the model for optimal performance combined with a simple two-stage pretraining setup.

Due to the limited scope of many of the datasets in our combined dataset, our models undergo an initial pretraining step using ImageNet. Most RGB cameras do not provide wavelength and bandwidth information. To deal with this, we utilize a dataset \cite{jiang2013space} of spectral sensitivity curves of 28 common DSLR cameras to compute an average wavelength and bandwidth for each channel in standard RGB images. These values are then passed into $L_{\lambda}$ and $L_{B}$ for all ImageNet images. Models are pretrained for 100 epochs on ImageNet before a further pretraining stage of 100 epoch using our hyperspectral imaging dataset. 

Training uses the AdamW \cite{loshchilov2017decoupled} optimizer with identical parameters to the original MAE paper. We used a cosine annealing with warm restart loss schedule, which resets the loss every 40 epochs. ImageNet pretraining is done with a batch size of 512, and hyperspectral pretraining uses a batch size of 16. As the masking is a sufficient regularizer \cite{he2022masked}, only vertical and horizontal flipping and RandomResizedCrop augmentations are used.
\section{Experiments}
\label{sec:experiments}

We evaluate our experiments on several downstream tasks, including both proximal and remote sensing datasets. For proximal datasets, we evaluate on the DriedFoods and Wheat datasets, and for remote sensing, we evaluate on the BigEarthNet \cite{clasen2024reben} dataset. 

\subsection{Finetuning procedure}

For the dried foods and wheat datasets, the input images to our fine-tuning models contain 50 evenly spaced channels. For the BigEarthNet dataset, we train using only 5\% of the training and validation data utilizing all 12 channels. We evaluate the BigEarthNet dataset on the entire test set.

All models are trained for 100 epochs on the finetuning dataset unless otherwise stated. The best-performing model on the validation set is saved and evaluated on a separate test set. Data augmentations include random horizontal and vertical flipping, CutMix \cite{yun2019cutmix}, and Mixup \cite{zhang2017mixup}.

\subsection{Channel count}
The number of channels used during pretraining changes the number of tokens in the sequence that is fed into the transformer. Here, we evaluate 4 models, pretrained on our dataset with 25, 50, 75, and 100 channel images with a batch size of 16. During pretraining, the image channels, as previously described, are chosen at random from the full hyperspectral image. To ensure each model sees the same number of pixels, the models are trained for a number of epochs based on the number of channels per image. For example, the 25-channel model is pretrained for 400 epochs, while the 100-channel model is pretrained for 100 epochs. For all finetuning tasks, 50 channels are selected at a uniform spacing from the image as input. Results are displayed in \cref{tab:channel}. Reconstructions can be seen in \cref{fig:recon}.

\begin{table}[ht]
\begin{tabular}{cccc}
\hline
Channels & DriedFoods & Wheat & BigEarthNet \\ \hline
25 & \textbf{93.18} & \textbf{66.49} & \textbf{69.34} \\
50 & 85.82 & 60.67 & 68.73 \\
75 & 83.18 & 61.90 & 68.95 \\
100 & 81.02 & 58.96 & 67.47 \\ \hline
\end{tabular}
\caption{F1 scores for the number of channels used during pretraining.}
\label{tab:channel}
\end{table}

We attribute the large difference in performance among these models to the amount of training the model has received. Due to the way we have implemented the pretraining, the 25-channel model has seen 4 times as many weight updates as the 100-channel model. Additionally, due to the RandomResizedCrop augmentation it has seen a large number of different images.

\subsection{Masking ratio} 
We know from \cite{he2022masked} that the optimal masking ratio for RGB imaging is 75\%. For video, this ratio increases all the way to 90\% \cite{tong2022videomae,feichtenhofer2022masked}. Compared to video, the spectral redundancy in hyperspectral images is higher than the temporal redundancy in video. This suggests a higher masking ratio should be used. To do so we evaluate our two-step masking process by modifying only the spectral masking ratio from 65\% to 95\% at 10\% intervals, resulting in a total masking ratio between 91.25\% and 98.75\%. The spatial masking ratio remains constant at 75\%. These models are all pretrained utilizing images generated from 100 randomly selected channels with a batch size of 16. Results are displayed in \cref{tab:masking} and reconstructions in \cref{fig:recon}.

\begin{table}[ht]
\begin{tabular}{cccc}
\hline
Ratio & DriedFoods & Wheat & BigEarthNet \\ \hline
65\% (91.25\%) & 80.66 & 56.17 & 67.44 \\
75\% (93.75\%) & 81.02 & 58.96 & 67.47 \\
85\% (96.25\%) & 84.54 & \textbf{61.59} & \textbf{68.07} \\
95\% (98.75\%) & \textbf{85.72} & 58.62 & 67.61 \\ \hline
\end{tabular}
\caption{F1 scores for the spectral masking ratio (total masking ratio) used during pretraining.}
\label{tab:masking}
\end{table}

The optimal masking ratio is around 85\%, suggesting a very high number of pixels (96.25\%) can be dropped to achieve good results. Higher masking ratios also improve pretraining time by reducing the number of patches that the larger ViT encoder must compute for each sample.

\subsection{Baseline experiments}
Here we evaluate using a few common deep learning methods in order to compare performance with our model. We use both random initialization and ImageNet-1K pretrained variants of ResNet-50~\cite{He_2016_CVPR} and ViT-Large \cite{dosovitskiy2020image}. During training 50 evenly spaced channels are selected and then compressed down to 3 channels with a single $3\times3$ convolutional layer in order to fit with the predefined 2D architectures. Results are displayed in \cref{tab:baselines}.

\begin{table}[ht]
\begin{tabular}{@{}cccc@{}}
\toprule
Baseline & DriedFoods & Wheat & BigEarthNet \\ \midrule
ResNet-50 & 18.42 & 16.08 & 68.36 \\
ResNet-50 (IN-1K) & 91.21 & 46.65 & 67.89 \\
ViT-L & 0.00 & 33.34 & 0.00 \\
ViT-L (IN-1K) & \textbf{97.98} & 56.17 & \textbf{70.96} \\ \midrule
Ours & 93.18 & \textbf{66.49} & 69.34 \\ \bottomrule
\end{tabular}
\caption{F1 scores for baseline ResNet-50 and ViT models.}
\label{tab:baselines}
\end{table}

The randomly initialized ViT fails to converge for two of the three datasets after 100 epochs, while the pretrained ViT outperforms all other baselines. Randomly initialized ResNet-50 performs worse than the pretrained variants in two out of the three datasets.

\subsection{Longer pretraining}
We also pretrained a number of additional models to evaluate the performance of the model when trained for more epochs, larger batch sizes, and higher masking ratios. Some are continued training where the 25 channel, 75\% masking ratio model left off in the previous experiments, while others begin immediately after ImageNet pretraining. These models are pretrained for longer without utilizing warm restarts, and instead opting for a standard cosine annealing schedule as in the MAE paper. Results are displayed in \cref{tab:longer}.

\begin{table*}[ht]
\centering
\begin{tabular}{@{}ccclcccc@{}}
\toprule
Epochs & Batch Size & Continued & Channels & Mask Ratio & DriedFoods & Wheat & BigEarthNet \\ \midrule
913 & 16 & True & 25 & 75\% & 92.58 & \textbf{66.48} & \textbf{69.90} \\
151 & 128 & True & 25 & 85\% & \textbf{94.48} & 64.96 & -- \\
277 & 16 & False & 25 & 85\% & 82.60 & 62.92 & 67.87 \\
166 & 128 & False & 25 & 85\% & 92.95 & 64.35 & 67.92 \\
162 & 1024 & False & 25 & 85\% & 92.54 & 59.64 & 68.61 \\ \bottomrule
\end{tabular}
\caption{Results for longer training. Continued refers to whether or not the model is resumed from the state saved during the first 400 epochs of the 25-channel, 75\% masking ratio model from the previous experiments.}
\label{tab:longer}
\end{table*}

According to the results from the table displayed here, training for longer or with larger batch sizes doesn't seem to improve the results significantly. The largest difference occurs when evaluating finetuning performance on the DriedFoods dataset using a model trained with a larger batch size (128) and higher masking ratio (85\%). There we see an F1 score of 94.48 as compared to our previous best result of 93.18 in \cref{tab:channel}.

\begin{figure*}[ht]
     \centering
     \includegraphics[width=0.99\linewidth]{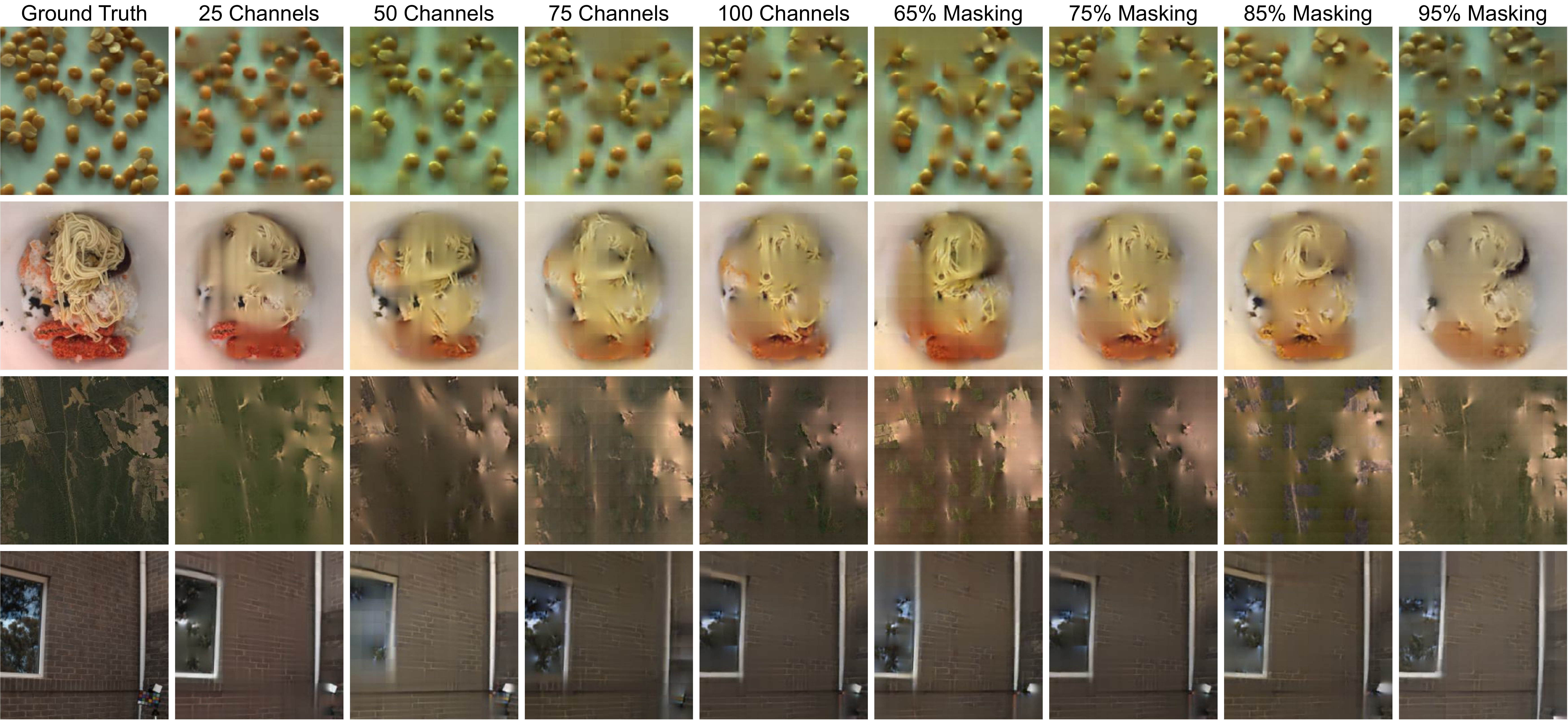}
     \caption{Reconstruction samples from the channel count and mask ratio experiments in RGB.}
     \label{fig:recon}
\end{figure*}
\section{Discussion}
\label{sec:discussion}

From \cref{tab:channel}, we see that increasing the number of channels seems to result in worse performance. We train the models with fewer channels for more epochs to ensure that each model sees the same number of pixels. This results in the 25-channel model being trained for 4 times as many epochs as the 100-channel model. In \cref{tab:datasets}, we see that most of the images in our dataset are $512 \times 512$ or larger and the RandomResizedCrop augmentations select $224 \times 224$ regions from each image. Therefore, a single epoch will not see all possible pixels in the dataset. This suggests that longer training improves performance due to a larger number of unique pixels seen. Further evaluation is necessary, but this can be tested by training the 100-channel model for 400 epochs and comparing the results. Additionally, the 25-channel model is the lowest number of channels that we tested, and its possible that even fewer channels trained for longer could perform even better.

Despite this, it seems training for longer than 400 epochs on the 25-channel model does not seem to increase performance by any significant amount. We only see an improvement in DriedFoods when training longer. We may be close to the upper limit of what our architecture can achieve with the given pretrained dataset. However, further investigation is needed to verify this.

Increasing the masking ratio also seems to improve model performance on downstream tasks. On all three downstream tasks, the best performance occurs with an 85\% or 95\% channel masking ratio. Due to the redundancy along the spectral dimension, it makes sense that a higher masking ratio performs better because a lot of the spectral information is unnecessary. Higher masking ratios are also beneficial to pretraining speed.

Visually, it is difficult to distinguish differences from the reconstructed images in \cref{fig:recon}. The 25-channel model seems to have the best color accuracy in most of the sampled images, while the 50-channel model seems to be the worst in color accuracy. The minor differences between these are likely dependent on which channels were dropped during the the spectral masking stage.

Compared to the baseline ResNet-50 and ViT-L models, our pretrained model outperforms all baselines except the ViT-L pretrained on ImageNet-1K. However, on the Wheat dataset, our model significantly outperforms the pretrained ViT baseline. We expect that, in the other two datasets (DriedFoods and BigEarthNet), the important spectral information has a stronger signal that can be retrieved by the singular convolutional layer at the beginning of the baseline architectures. For example, the DriedFoods dataset consists of objects that vary significantly in shape and color suggesting that we could quite easily classify with RGB images alone. In contrast, the important information for classifying wheat plant age might be contained within varying shades of green. Our channel-wise model may be able to better pick up these more complex, weaker signals.

Another interesting factor is that our best-performing model (25 channels, 75\% channel masking ratio, in \cref{tab:channel}) is pretrained on 25 channel images and finetuned on 50 channel images suggesting that our implementation of spatial-spectral positional encoding is robust to changes in sequence length along the spectral dimension. In fact, this seems reasonable as the way our implementation works is that we are actually doing positional interpolation along the spectral dimension rather than extrapolation as discussed in \citet{posint}. This happens because our spectral encoding is uniquely defined per sample. It is not the same encoding for each sample and instead is dependent on wavelength and bandwidth of the channels in the sample. Each sample is a random selection of channels from the original image and the dataset contains a number of different cameras. This results in a spectral encoding that is very different between each sample. This is an interesting property that could benefit from further investigation. 

Having a unique positional encoding for each sample could also be a detriment to model performance. One of the reasons we utilize positional encodings when working with transformer architectures is that they lack any inherent positional knowledge. It may be the case that having a unique positional encoding for each image is too difficult for the model to converge optimally.
\section{Conclusion}
\label{sec:conclusion}

In conclusion, we have built the first hyperspectral foundational model that incorporates proximal spectral imaging by utilizing positional encoding of the spectral channels and the transformer-based masked auto-encoder (MAE) to model input images with varying numbers of spectral channels at varying wavelengths. Hereby, we are able to utilize the small proximal datasets as opposed to only including more widely available remote sensing imagery. Our model outperforms standard baselines on proximal imaging datasets that require complex spectral knowledge and is robust to variations in sequence length. It is built on the MAE architecture, with a unique two-stage masking method designed to ensure that both spatial and spectral knowledge is learned and a spatial-spectral encoding that utilizes wavelength and bandwidth information.
{
    \small
    \bibliographystyle{ieeenat_fullname}
    \bibliography{main}
}


\end{document}